\documentclass[10pt,twoside,frenchb]{article}

\usepackage{graphicx}

\usepackage{epsf}

\usepackage{amssymb}
\usepackage{amsmath}
\usepackage{latexsym}
\usepackage{comment}
\usepackage{graphicx}
\usepackage{comment}
\usepackage{textcomp}
\usepackage{float}
\usepackage[locale=FR]{siunitx}
\usepackage{subfigure}
\usepackage{setspace}
\usepackage{comment}
\usepackage{hyperref}
\usepackage{subfiles}

\usepackage{taln2019}
\usepackage{babel}

\usepackage{tikz}
\usetikzlibrary{calc}
\usetikzlibrary{positioning}
\tikzset{
    >=stealth,
    hair lines/.style={line width = 0.05pt, lightgray},
    true scale/.style={scale=#1, every node/.style={transform shape}},
}

\DeclareMathOperator\FFNN{FFNN}

\DeclareMathOperator\logsoftmax{log-softmax}
\DeclareMathOperator\logP{log-P}
\DeclareMathOperator\argmax{argmax}
\DeclareMathOperator\GRU{GRU}

\DeclareMathOperator\fGRU{\overrightarrow{\GRU}}
\DeclareMathOperator\bGRU{\overleftarrow{\GRU}}
\DeclareMathOperator\Norm{Norm}

\title{Hybrid Neural Models For Sequence Modelling: The Best Of Three Worlds}

\author{Marco Dinarelli$^{1}$\quad Lo\"{i}c Grobol$^{2,3}$\\
  {\small
    (1) LIG, B\^atiment IMAG - 700 avenue Centrale - Domaine Universitaire de Saint-Martin-d’H\`eres \\ 
    (2) Lattice CNRS, 1 rue Maurice Arnoux, 92120 Montrouge, France \\ 
    (3) ALMAnaCH Inria, 2 rue Simone Iff, 75589 Paris, France \\ 
    \texttt{
      marco.dinarelli@univ-grenoble-alpes.fr, loic.grobol@gmail.com \\ 
}}}

\begin{document}
\maketitle

\abstract{}{
\vspace{-1.0em}
  We propose a neural architecture with the main characteristics of the most successful neural models of the last years: bidirectional RNNs, \emph{encoder-decoder}, and the \emph{Transformer} model. Evaluation on three sequence labelling tasks yields results that are close to the state-of-the-art for all tasks and better than it for some of them, showing the pertinence of this hybrid architecture for this kind of tasks.
\vspace{0.5em}
}


\section{Introduction}

Sequence labelling is an important NLP task, many NLP problems can be modeled indeed as sequence labelling.
The most traditional labelling problems are Part-Of-Speech (POS) tagging, syntactic chunking, Named Entity Recognition (NER) \cite{Collobert-2011-NLP-1953048.2078186}, and also Spoken Language Understanding (SLU) in the context of human-machine dialog systems \cite{demori08-SPM}.

Moreover there are other NLP problems that can be decomposed into several processing steps, the first of which can be often modeled as a sequence labelling problem.

We can place among such problems the syntactic constituency parsing, which can be decomposed into POS tagging and constituents analysis \cite{Collins-1997-TGL}~; the coreference chain resolution \cite{Soon2001,Ng2002}, composed of mention recognition from raw text and coreferent mention detection~; and also the structured named entity recognition \cite{Grouin.etAL:I2B2:2011,dinarelli2012-eacl,DINARELLI-ROSSET:OCR-NER:LREC2012}.

More widely, machine translation and syntactic constituency parsing can be modeled as end-to-end sequence prediction problems \cite{Sutskever-2014-SSL-2969033.2969173,DBLP-journals-corr-BahdanauCB14,46201,DBLP-journals-corr-VinyalsKKPSH14}, as well as a wide class of language understanding tasks \cite{devlin2018BERTPretrainingDeep}.
It would be possible thus, at least in principle, to design a model for sequence prediction in a multi-task learning framework, in order to deal with most of the NLP tasks, which shows in turn the interest in the research for sequence modelling problems.

The NLP problems that can be decomposed into several processing steps can be solved also with a single model.
For example the syntactic constituency parsing can be performed end-to-end with a model performing POS tagging and constituent analysis at the same time, e.g. \cite{Rush-2012-IPP-2390948.2391112}.
However, even in such cases, pre-training representations with neural networks and fine-tuning them afterwards on simpler tasks, like sequence labelling, can still improve the prediction performance \cite{peters2018DeepContextualizedWord}.
We can also consider this training procedure similar to embedding pre-training, which effectiveness has already beed proved  \cite{lample2016neural,Ma-Hovy-ACL-2016}.

In this paper we focus on neural architectures for sequence modelling in the sense this is meant in problems such as POS tagging, morpho-syntactic analysis, and semantic chunking and tagging. The latter is performed for SLU tasks in the context of human-machine dialog systems \cite{demori08-SPM}.

We inspired our work from \cite{DBLP-journals-corr-abs-1804-09849}, from which we also inspired the title of our paper, and which proposes neural hybrid models combining the \emph{Encoder-Decoder} (or \emph{Sequence-to-Sequence}) and \emph{Transformer} architectures. We propose an architecture with the main characteristics of the most successful neural models of the last few years: the bidirectional RNNs \cite{lample2016neural,Ma-Hovy-ACL-2016}, the \emph{Sequence-to-Sequence} model \cite{Sutskever-2014-SSL-2969033.2969173,DBLP-journals-corr-BahdanauCB14}, and the \emph{Transformer} \cite{46201}.

We evaluate our architecture on three traditional tasks of sequence labelling: Spoken Language Understanding (SLU) in French (MEDIA corpus) \cite{Bonneau-Maynard2006-media}, POS tagging in English (WSJ corpus) \cite{Marcus93buildinga}, and morpho-syntactic tagging in German (TIGER corpus) \cite{Brants_2004}.
The results outperform often the state-of-the-art, and allow in any case to conclude that our architecture finds its place among neural architectures.

\section{Neural Architectures}
\label{sec:neural-models}

The proposed neural models are inspired from the \emph{LSTM+CRF} architectures \cite{lample2016neural,Ma-Hovy-ACL-2016} in the way input encoding (tokens, characters, etc.) is performed. They are inspired from the \emph{Sequence-to-Sequence} architecture for the overall architecture, and from the models proposed in our previous work \cite{2016:arXiv:DinarelliTellier:NewRNN,DinarelliTellier:RNN:CICling2016,dinarelli_hal-01553830,Dupont-etAl-LDRNN-CICling2017,DBLP-journals/corr/abs-1904-04733} for making predictions based on a bidirectional context on the output side (labels).
Afterwards we added to this architecture some of the characteristics of the \emph{Transformer} model.

This work constitutes a step ahead with respect to \cite{DBLP-journals/corr/abs-1904-04733}. While we have not run full experiments to show the improvements with respect to our previous work, and results are equivalent in some cases, some outcomes suggest that the model proposed here is more effective. In any case the model described here is more general, as the architecture has been slightly modified so that it can be applied also to tasks like machine translation and syntactic parsing, though we do not evaluate on these tasks in this work. Moreover in other cases results are state-of-the-art, though they are obtained on data (TIGER) on which we did not evaluate previously.

\subsection{Encoder}
\label{subsec:encoder}

The encoder of our network is made of a bidirectional GRU \cite{Cho-2014-GatedRecurrentUnits}, which takes as input a sequence $S^{lex}$ representing input tokens as the concatenation of word embeddings $\left(E_w(w_i)\right)$ and representations computed from their characters $h_c(w_i)$. The latter is computed by an \emph{ad-hoc} recurrent layer, in a fashion similar to those used in the literature \citet{Ma-Hovy-ACL-2016}, with the difference that we use a GRU recurrent layer instead of a convolution layer.

The character-level representation of a word $w$ is computed like in \cite{DBLP-journals/corr/abs-1904-04733}:

\begin{equation}\label{eq:charrep}
    \begin{aligned}
        S^c(w) &= (E_{c}(c_{w, 1}), \dots, E_c(c_{w, n})) \\
        (h_c(c_{w, 1}), \dots, h_c(c_{w, n})) &= \GRU_c(S^c(w), h_c^0) \\
        h_c(w) &= \FFNN( Sum( h_c(c_{w, 1}), \dots, h_c(c_{w, n}) ) )
    \end{aligned}
\end{equation}

where $E_c$ is the matrix of character embeddings, $c_{w,i}$ is the \emph{i-th} character of the word $w$, $S^c(w)$ is the sequence of character embeddings for the word $w$, $\GRU_c$ is the GRU layer for characters, $h_c(c_{w, i})$ is the hidden state associated to the \emph{i-th} character of the word $w$. $\GRU_c$, like $\GRU_w$, is a bidirectional GRU layer.

The full representation of the \emph{i-th} word $w_i$ in a sequence is computed as:

\begin{equation}\label{eq:lex-rep}
    \begin{aligned}
        S^{lex} &= ([E_{w}(w_{1}), h_{c}(w_1)], \dots, [E_{w}(w_{N}), h_{c}(w_N)]) \\
        (h_{w_1}, \dots, h_{w_N}) &= \GRU_w(S^{lex}, h_w^0)
    \end{aligned}
\end{equation}

Since $\GRU_w$ scans the sequence forward and backward, $h_{w_i}$ depends on the word $w_i$ and its context.
When additional features are available at token level, they are embedded in the same way as words and concatenated to word embeddings in the sequence $S^{lex}$.

\subsection{Decoders}
\label{subsec:decoder}

Our model uses a representation of both left and right context at label level like proposed in \citet{2016:arXiv:DinarelliTellier:NewRNN,dinarelli_hal-01553830,Dupont-etAl-LDRNN-CICling2017,DBLP-journals/corr/abs-1904-04733}.

We use a $\bGRU_e$ \emph{backward} layer for encoding the right context, and a $\fGRU_e$ \emph{forward} layer for the left context with respect to a given labelling position $i$ in the input sequence.
These layers take as input the lexical level representation computed by the encoder and the label embeddings $E_e(e_i)$, which makes them similar to the decoder used in the original \emph{Sequence-to-Sequence} architecture for machine translation \citet{Sutskever-2014-SSL-2969033.2969173,DBLP-journals-corr-BahdanauCB14}.
An evolution with respect to such architecture is the use of two decoders, one for the left label-level context and one for the right label-level context.
This idea of using both left and right context at the output item's level has attracted some interest in the last few years, and has been applied in different variants to machine translation \cite{NIPS2017_6775,DBLP-journals/corr/abs-1801-05122}.

The right label-level context is computed by the \emph{backward} $\bGRU_e$ decoder as follows:

\begin{equation}\label{eq:label-bw-le}   
    \overleftarrow{h_{e_i}} = \bGRU_e([h_{w_i}, E_e(e_{i+1})], \overleftarrow{h_{e_{i+1}}})
\end{equation}

The left label-level context $\overrightarrow{h_{e_i}}$ is computed in a similar way by the \emph{forward} $\fGRU_e$ decoder.

\subsection{Output Layer}
\label{subsec:output}

In order to allow the model to make global decisions when predicting the current label, we add an output layer on top of the backward decoder, made of an affine transformation followed by a \emph{log-softmax} activation. This computes log-probabilities of backward predictions: 

\begin{equation}\label{eq:backward-model}
    \begin{aligned}
        \logP(\overleftarrow{e_{i}}) &= \logsoftmax( W_{bw} [h_{w_{i}}, \overleftarrow{h_{e_i}}] + b_{bw} ) \\
        \overleftarrow{e_{i}} &= \argmax( \logP(\overleftarrow{e_{i}}) )
    \end{aligned}
\end{equation}

The forward decoder predicts labels using as input the label-level context representations $\overrightarrow{h_{e_i}}$ and $\overleftarrow{h_{e_i}}$, as well as the lexical level information $h_{w_i}$ computed by the encoder:

\begin{equation}\label{eq:bidirectional-model}
    \begin{aligned}
        \logP(\overrightarrow{e_{i}}) &= \logsoftmax( W_o [\overrightarrow{h_{e_i}}, h_{w_i}, \overleftarrow{h_{e_i}}] + b_o ) \\
        \overrightarrow{e_i} &= \argmax( \logP(\overrightarrow{e_{i}}) )
    \end{aligned}
\end{equation}

In order to strengthen the global character of the decision, the final output of the model is computed as the arithmetic mean (in the log space, which gives a geometric mean of probabilities) of the \emph{forward} and \emph{backward} log-probabilities: $\frac{1}{2} ( \logP(\overrightarrow{e_{i}}) + \logP(\overleftarrow{e_{i}}) )$.
This \emph{teaches} the model to compute real good predictions also in the backward phase, instead of basing decisions on raw heuristics, or even on the forward (left) context only. Indeed, the backward decoder has only a partial view of the output level information, while the forward decoder uses both left and right context.

\subsection{Learning}
\label{subsec:learning}

All models described in this paper are learned minimizing the minus log-likelihood $\mathcal{LL}$ of the training data. More precisely:

\begin{equation}\label{eq:LL}
    -\mathcal{LL}(\Theta | D) = -\sum_{d=1}^{|D|} \sum_{i=1}^{N_d} \frac{1}{2}(\logP(\overrightarrow{e_i})+\logP(\overleftarrow{e_i}))  + \frac{\lambda}{2} \left | \Theta \right |^2 )
\end{equation}

The first sum scans the data $D$ of size $|D|$, while the second sum is over training examples $S_d$, of size $N_d$.

Since data used in this work for evaluation are relatively small, while our models are relatively complex, we add a Gaussian regularization term $L_2$ with a coefficient $\lambda$.

\subsection{\emph{The best of three worlds}}
\label{subsec:bestof3worlds}

The neural architecture described so far uses the same ideas introduced in \cite{2016:arXiv:DinarelliTellier:NewRNN,dinarelli_hal-01553830,Dupont-etAl-LDRNN-CICling2017,DBLP-journals/corr/abs-1904-04733} for predicting labels using both representations of left (forward) and right (backward) contexts, and for both input-level information (words, characters, etc.) and labels.
Our model integrates also the characteristics of bidirectional RNNs and of Encoder-Decoder architectures. Moreover we use two decoders, instead of one like in the original architecture.

Starting from this neural architecture we added also some of the characteristics of the \emph{Transformer} model.

We note that the paper from which we inspired our work \cite{DBLP-journals-corr-abs-1804-09849} investigated the combination of two Encoder-Decoder architectures, one based on recurrent layers \cite{Sutskever-2014-SSL-2969033.2969173,DBLP-journals-corr-BahdanauCB14}, while the other is based on the \emph{Transformer} model, which is not a recurrent network.
This work shows that such combination is possible and effective compared to the two combined architectures.

In this work we combined three different architectures with the goal of putting together their strengths, and trying to overcome some limitations.

First of all, in the tasks of sequence labelling used for evaluation in this work, there is a \emph{one-to-one} correspondence between input and output items (words and labels, respectively). Our preliminary experiments on mention detection for coreference resolution using a \emph{Transformer}, as well as some experiments in the literature \cite{DBLP-journals/corr/abs-1902-09113}, suggest that ignoring such correspondence leads to a remarkable performance drop.\footnote{F1 measure drops more than 3 points using the \emph{Transformer} in our experiments}
For this reason the encoder and the decoders used in our architecture exploit the one-to-one correspondence of sequence labelling tasks. We are aware though this information could not be used in \emph{real} sequence-to-sequence problems such like machine translation or syntactic parsing.

Second, we chose an Encoder-Decoder overall architecture since this proved to be the most effective for sequence-to-sequence problems (including sequence labelling). Moreover, we showed that keeping a label-level context into account, in some sequence labelling problems, using a decoder is more effective than doing the same with a CRF neural layer \cite{dinarelli_hal-01553830}, we will show the same also in this work.
As we already mentioned, in our architecture we use two decoders. This solution allows for an even more effective modelling of a label-level context. This solution is not possible, at least in principle, with the original \emph{Transformer} model.

Third, we decided to integrate some of the characteristics of the \emph{Transformer} in order to overcome limitations of RNNs concerning the learning phase. As discussed in \cite{46201}, RNNs are difficult to train because of the length of paths the learning signal has to back-propagate through. The length of such paths is the consequence of the recurrent nature of the networks.

The \emph{Transformer} model \cite{46201} proposed an alternative, non-recurrent Encoder-Decoder architecture based on a multi-head attention mechanism.
Recent works \cite{Dehghani2018UniversalT,dai2019TransformerXLAttentiveLanguage} suggest however that using some kind of recurrent connections \emph{Transformer}'s performance can be improved.

Another interesting characteristic of the \emph{Transformer} is the use of residual connections, or skip-connections \cite{Bengio03aneural}.
This characteristic allows for alleviating the vanishing gradient problem of deep networks.
Indeed, the use of gates in recurrent networks reduces the amplitude of this problem, however the literature and our own experiments suggest that for long sequences, gates are not sufficient for an effective learning.

The attention mechanism does not suffer from this problem. Each element of the input sequence is related to a small, fixed number of layers in the attention mechanism. The back-propagation path of the learning signal is thus much shorter than that of RNNs.
Moreover the use of skip-connections contributes in keeping a higher learning signal amplitude, allowing to skip a whole layer, over which the learning signal would be weakened (because of gradient multiplications).

\begin{figure}
\center
\begin{tikzpicture}
	
	\begin{scope}[local bounding box=net]
	
	\node (o) at (0,5) {output};
	\node (oTrans) at (-5,5) {output};
	\node (oSeq2Biseq) at (5,5) {output};
	
	\node[draw, rectangle, rounded corners=3pt, fill=gray!50] (a2) at (0,4) {\emph{Add\&Norm}};
	\node[draw, rectangle, rounded corners=3pt, fill=yellow!75] (a2Trans) at (-5,4) {\emph{Add\&Norm}};
	\node[draw, rectangle, rounded corners=3pt, fill=yellow!75] (a2Seq2Biseq) at (5,4) {\emph{Add\&Norm}};
	
	\node[draw, rectangle, rounded corners=1pt, scale=1.2] (t) at (0,3) {\emph{deep-encoder}};
	\node[draw, rectangle, rounded corners=1pt, scale=1.2, fill=cyan] (tTrans) at (-5,3) {\emph{FFNN}};
	\node[draw, rectangle, rounded corners=1pt, scale=1.2, fill=cyan] (tSeq2Biseq) at (5,3) {\emph{FFNN}};
	
	\node[draw, rectangle, rounded corners=3pt, fill=gray!50] (a1) at (0,2) {\emph{Add\&Norm}};
	\node[draw, rectangle, rounded corners=3pt, fill=yellow!75] (a1Trans) at (-5,2) {\emph{Add\&Norm}};
	\node[draw, rectangle, rounded corners=3pt, fill=yellow!75] (a1Seq2Biseq) at (5,2) {\emph{Add\&Norm}};
	
	\node[draw, rectangle, rounded corners=1pt, scale=1.2] (c) at (0,1) {\emph{ctx-encoder}};
	\node[draw, rectangle, text width=2cm, text centered, rounded corners=1pt, scale=1.2, fill=orange] (cTrans) at (-5,1) {\emph{Multi-head Attention}};
	\node[draw, rectangle, rounded corners=1pt, scale=1.2, fill=orange] (cSeq2Biseq) at (5,1) {\emph{GRU}};
	
    \node (i) at (0,-0.3) {input};
    \node (iTrans) at (-5,-0.3) {input};
    \node (iSeq2Biseq) at (5,-0.3) {input};
    
    \draw[->, dashed] (i) -- (c);
    \draw[->, dashed] (iTrans) -- (cTrans);
    \draw[->, dashed] (iSeq2Biseq) -- (cSeq2Biseq);
    
    \draw[->] (c) -- (a1);
    \draw[->] (cTrans) -- (a1Trans);
    \draw[->] (cSeq2Biseq) -- (a1Seq2Biseq);
    
    \draw[->] (a1) -- (t);
    \draw[->] (a1Trans) -- (tTrans);
    \draw[->] (a1Seq2Biseq) -- (tSeq2Biseq);
    
    \draw[->] (t) -- (a2);
    \draw[->] (tTrans) -- (a2Trans);
    \draw[->] (tSeq2Biseq) -- (a2Seq2Biseq);
    
    \draw[->, dashed] (a2) -- (o);
    \draw[->, dashed] (a2Trans) -- (oTrans);
    \draw[->, dashed] (a2Seq2Biseq) -- (oSeq2Biseq);
    
    \draw[->, red, thick] (i) to[bend left=90] (a1);
    \draw[->, thick, red] (iTrans) to[bend left=90] (a1Trans.west);
    \draw[->, thick, red] (iSeq2Biseq) to[bend left=90] (a1Seq2Biseq);

    \draw[->, thick, red] (a1.north west) to[bend left=90] (a2.south west);
    \draw[->, thick, red] (a1Trans.north west) to[bend left=90] (a2Trans.south west);
    \draw[->, thick, red] (a1Seq2Biseq.north west) to[bend left=90] (a2Seq2Biseq.south west);

    \tikzstyle{noeud}=[minimum width=3.2cm,minimum height=4.0cm, rectangle,rounded corners=10pt,draw,text=red,font=\bfseries]
    \node[noeud] (N) at (0,2.5) {};

    \tikzstyle{noeud}=[minimum width=4.2cm,minimum height=4.3cm, rectangle,rounded corners=10pt,draw,text=red,font=\bfseries]
    \node[noeud] (NTrans) at (-5,2.3) {};
    
    \tikzstyle{noeud}=[minimum width=3.2cm,minimum height=4.0cm, rectangle,rounded corners=10pt,draw,text=red,font=\bfseries]
    \node[noeud] (NSeq2Biseq) at (5,2.5) {};

    \end{scope}

\end{tikzpicture}
    \caption{High-level architecture, from a conceptual point-of-view, of each block of a \emph{Transformer} model (in the middle). On the left we show its instantiation in the \emph{Transformer}, on the right we show its instantiation in our architecture, where we use a bidirectional GRU layer to encode contextual features}\label{fig:generaltransformer}
\end{figure}
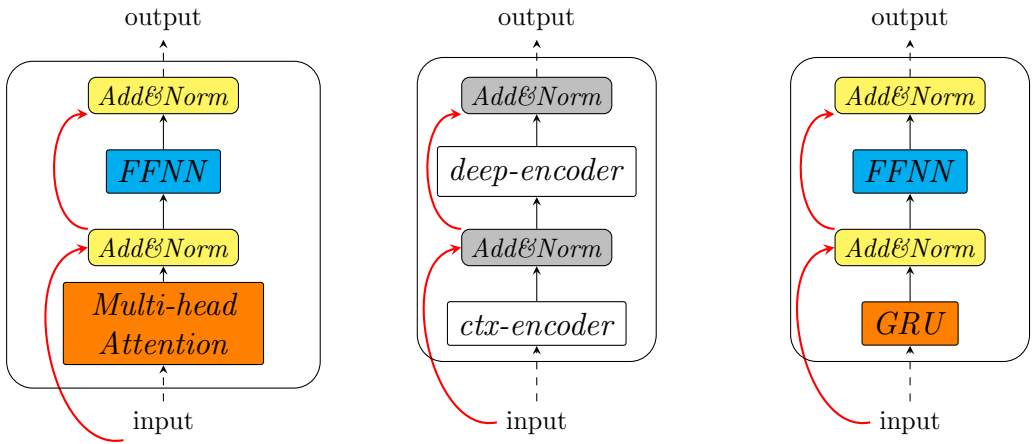

We can interpret each block of a \emph{Transformer} architecture, both encoder and decoder, as the combination of a sub-module which \emph{``encodes''} contextual features, we call this sub-module \emph{ctx-encoder}, and a \emph{feed-forward} sub-module which \emph{``maps''} contextual features into a deeper feature space, we call the second sub-module \emph{deep-encoder}.
The output of the two sub-modules is added to their respective input (skip-connection) and normalized with a \emph{layer normalization}.
This interpretation is shown in the figure~\ref{fig:generaltransformer} in the middle. Skip-connections are highlighted in red.

The \emph{Transformer} proposed in \citet{46201} instantiate the generic architecture described above using as \emph{ctx-encoder} the multi-head attention mechanism, while it uses a two-layer feed-forward network as \emph{deep-encoder}. A single block of this architecture is shown in the figure~\ref{fig:generaltransformer} on the left.

We modified our neural architecture for instantiating the generic architecture. In our solution we use a bidirectional GRU layer as \emph{ctx-encoder}, while we use exactly the same sub-module as the \emph{Transformer} as \emph{deep-encoder}. A block of our architecture is shown in the figure~\ref{fig:generaltransformer} on the right.

The characteristics of the \emph{Transformer} model we integrated in our architecture are thus the skip-connections and layer normalization (\emph{Add\&Norm} in the figure~\ref{fig:generaltransformer}), as well as the two-layer feed-forward sub-module to re-encode contextual features once the skip-connection has been applied (\emph{FFNN} in the figure~\ref{fig:generaltransformer}).
Following the computation chain suggested by \citet{DBLP-journals-corr-abs-1804-09849}, the output of our encoder is computed as follows:

\begin{equation}\label{eq:lex-rep-trans}
    \begin{aligned}
        \hat{S}_i^{lex} &= \Norm(S_i^{lex}) \\
        \hat{h}_{w_i} &= GRU_w(\hat{S}_i^{lex}, h_{w_{i-1}}) \\
        h_{w_i} &= \FFNN(\Norm(\text{Dropout}(\hat{h}_{w_i}) + \hat{S}_i^{lex}))
    \end{aligned}
\end{equation}

\noindent where we use \emph{Norm} for the layer normalization, and \emph{Dropout} for a dropout regularization \cite{JMLR:v15:srivastava14a}.
The other GRU layers of our architecture, introduced in previous sections, are modified in a similar way.

In this work we don't use any attention mechanism like in the \emph{Transformer}, context is granted by the use of GRU layers (cf. section~\ref{subsec:encoder}).
Our choice is motivated by the work described in \citet{P18-2116}, where it is shown that RNNs, such like $\GRU$, can encode long contexts as a weighted mean of the inputs, in a way that can be considered similar to an attention mechanism.
Moreover, using $\GRU$ layers avoid the use of positional encoding, the sequential structure of an input sentence is encoded implicitly by the RNNs.\footnote{We don't exclude however that using positional encoding would improve our architecture}

\begin{figure}
\center
\begin{tikzpicture}
	\def\words{$w_1$, $w_2$, $w_3$}
	\def\foutputs{$\overrightarrow{e_1}$, $\overrightarrow{e_2}$, $\overrightarrow{e_3}$}
	\def\boutputs{$\overleftarrow{e_1}$, $\overleftarrow{e_2}$, $\overleftarrow{e_3}$}
	
	\begin{scope}[local bounding box=net]
	\foreach \w [count=\wi from 1, remember=\wi as \lastwi] in \words {
	\ifnum\wi>1
            	\node[right=6.5em of w\lastwi.center, text height=1.5ex, text depth=0.25ex, anchor=center] (w\wi) {\w};
            \else
            	\node[text height=1.5ex, text depth=0.25ex] (w\wi) {\w};
            \fi
	}
	\foreach \w [count=\wi from 1, remember=\wi as \lastwi] in \words {
		\draw[->] (w\wi) -- +(0, 2em) node[draw, anchor=south, rounded corners=1pt, inner sep=0.3em] (ew\wi) {encoder};
		\ifnum\wi>1
			\draw[<->] (ew\lastwi) -- (ew\wi);
		\fi
	}
	
	\foreach \w [count=\wi from 1, remember=\wi as \lastwi] in \words {
		\draw[->] (ew\wi) -- +(-0.5, 2em) node[draw, anchor=south, rounded corners=1pt, inner sep=0.3em] (bdec\wi) {$\overleftarrow{decoder}$};
		\ifnum\wi>1
			\draw[->] (bdec\wi) -- (bdec\lastwi);
		\fi
	}
	
	\foreach \w [count=\wi from 1, remember=\wi as \lastwi] in \words {
		\draw[->,red] (ew\wi) to[bend right=30] +(+0.5, 5em) node[draw, anchor=south, rounded corners=1pt, inner sep=0.3em] (fdec\wi) {$\overrightarrow{decoder}$};
		\draw[->] (bdec\wi) -- (fdec\wi);
		\ifnum\wi>1
			\draw[->] (fdec\lastwi) -- (fdec\wi);
		\fi
	}
	
	\foreach \y [count=\yi from 1, remember=\yi as \lastyi] in \foutputs {
            	\draw[red,->] (fdec\yi) -- +(0, 3em) node[text height=1.5ex, text depth=0.25ex, anchor=south] (fy\yi) {\y};
            	\ifnum\yi>1
            		\draw[red,->] (fy\lastyi) to[bend right=30] (fdec\yi);
            	\fi
	}
	
	\foreach \y [count=\yi from 1, remember=\yi as \lastyi] in \boutputs {
		\draw[->] (bdec\yi) -- +(0, 5.3em) node[text height=1.5ex, text depth=0.25ex, anchor=south] (by\yi) {\y};
	}
	
	\draw[->] (by3) to[bend left=42] (bdec2.east);
	\draw[->] (by2) to[bend left=42] (bdec1.east);

        \end{scope}

\end{tikzpicture}
    \caption{Overall architecture of our model, this is the same as the model presented in our previous work \citet{DBLP-journals/corr/abs-1904-04733}}\label{fig:network-structure}
\end{figure}
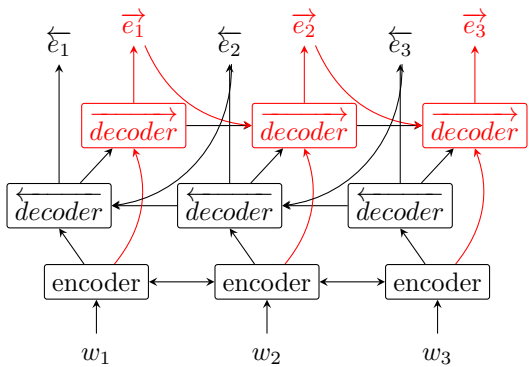

The overall architecture of our model is the same as our previous work \cite{DBLP-journals/corr/abs-1904-04733} (cf figure~\ref{fig:network-structure}), the difference is in the computation of layer's output described in previous sections.

\section{Evaluation}
\label{sec:eval}

We evaluate our models on three tasks of sequence labelling:

\textbf{The French corpus MEDIA}
\cite{Bonneau-Maynard2006-media} was created for evaluating spoken dialog systems in the context of hotel information and reservation in France.
Data have been annotated following a rich concept ontology. Simple semantic components can be combined to create more precise and complex semantic units.\footnote{For example the component \texttt{localization} (we translate from French) can be combined with \texttt{city}, \texttt{relative-distance}, \texttt{general-relative-localization}, \texttt{street}, etc.}
Concepts span in general more than one word, the goal is thus to tag words and then reconstruct semantic units from the tags.
Statistics of the corpus MEDIA are shown in figure~\ref{tab:MEDIAStats}.

This task is modelled as sequence labelling splitting semantic units over multiple words using the \textit{BIO} formalism \cite{Ramshaw95-BIO}. We already evaluated on this task in the past, using a large variety of probabilistic models \cite{dinarelli09:Interspeech,dinarelli09:emnlp,Quarteroni2009:Interspeech,Hahn.etAL-SLUJournal-2010,Dinarelli2010.PhDThesis,dinarelli2011:emnlp,Dinarelli.etAl-SLU-RR-2011}.

In the MEDIA corpus some word classes (clusters) are provided with the annotation. Such classes allow for a better generalization on words belonging to categories for which lists can be easily retrieved, or constructed, in the context of a real human-machine dialog system. Examples of such classes are the name of cities in France (class \emph{CITY}), the hotel brands (class \emph{HOTEL}), currency amounts (\emph{AMOUNTS}), etc. We used these classes in some of our experiments, this is indicated with \emph{FEAT} in the tables of the following section.

\begin{table}[t]
\begin{minipage}{1.0\linewidth}
    \centering
    \scriptsize
    \begin{tabular}{|l|rr|rr|rr|}
      \hline
      & \multicolumn{2}{|c|}{Training} & \multicolumn{2}{|c|}{Dev} & \multicolumn{2}{|c|}{Test}\\
      \hline
      \# sentences     &\multicolumn{2}{|c|}{12~908} &\multicolumn{2}{|c|}{1~259}&\multicolumn{2}{|c|}{3~005} \\
      \hline
      \hline
      & \multicolumn{1}{|c}{Words} & \multicolumn{1}{c|}{Concepts} &  \multicolumn{1}{|c}{Words} & \multicolumn{1}{c|}{Concepts} &
      \multicolumn{1}{|c}{Words} & \multicolumn{1}{c|}{Concepts} \\
      \hline
      \# tokens          & 94~466 & 43~078 & 10~849 & 4~705 & 25~606 & 11~383 \\
      dictionary         &  2~210 &     99 &    838 &    66 &  1~276 &     78 \\
      OOV\%   & --     & --     &  1,33  & 0,02  &  1,39  &  0,04  \\
      \hline
    \end{tabular}
    \caption{Statistics of the French corpus MEDIA}
  \label{tab:MEDIAStats}
  \end{minipage}
\end{table}

\textbf{The English Penn TreeBank corpus}
\cite{Marcus93buildinga}, named WSJ henceforth, is one of the most used corpus for evaluating models for sequence labelling. The task consists in assigning each word its POS tag.
For our experiments we use the most common data split: sections 0-18 are used as training data, sections 19-21 as development data, and sections 22-24 are used for testing.
Statistics of this corpus are given in the table~\ref{tab:WSJStats}.

\begin{table}[t]
\begin{minipage}{1.0\linewidth}
    \centering
    \scriptsize
    \begin{tabular}{|l|rr|rr|rr|}
      \hline
      & \multicolumn{2}{|c|}{Training} & \multicolumn{2}{|c|}{Dev} & \multicolumn{2}{|c|}{Test}\\
      \hline
      \# sentences     &\multicolumn{2}{|c|}{38~219} &\multicolumn{2}{|c|}{5~527}&\multicolumn{2}{|c|}{5~462} \\
      \hline
      \hline
      & \multicolumn{1}{|c}{Words} & \multicolumn{1}{c|}{Labels} &  \multicolumn{1}{|c}{Words} & \multicolumn{1}{c|}{Labels} &
      \multicolumn{1}{|c}{Words} & \multicolumn{1}{c|}{Labels} \\
      \hline
      \# tokens          & 912~344 & -- & 131~768 & -- & 129~654 & -- \\
      dictionary         &  43~210 &     45 &    15~081 &    45 & 13~968 &     45 \\
      OOV\%   & --     & --     & 22,61  & 0  &  20,00  &  0  \\
      \hline
    \end{tabular}
    \caption{Statistics of the English corpus WSJ}
  \label{tab:WSJStats}
  \end{minipage}
\end{table}

\textbf{The German corpusTIGER}
\cite{Brants_2004} is annotated with rich morpho-syntactic information. These include POS tags like in the WSJ corpus, but also gender, number, cases, and other inflection information, as well as conjugation information for verbs.
The combination of all these pieces of information constitutes the output labels.
This task is similar to the POS tagging from a sequence labelling point of view, however it is made more difficult on one side by the language (German has more complex syntactic structures than English), on the other side by the relatively large number of labels to be predicted (\num{694} in total, while there are \num{138} labels in MEDIA and \num{45} in the WSJ corpus).
We use the same data split used in \cite{Lavergne17learning}. Statistics of this corpus are given in the table~\ref{tab:TIGERStats}.
As we can see from tables, the size of dictionaries, for both words and labels, is relatively large in the WSJ and TIGER corpora.

\begin{table}[t]
\begin{minipage}{1.0\linewidth}
    \centering
    \scriptsize
    \begin{tabular}{|l|rr|rr|rr|}
      \hline
      & \multicolumn{2}{|c|}{Training} & \multicolumn{2}{|c|}{Dev} & \multicolumn{2}{|c|}{Test}\\
      \hline
      \# sentences     &\multicolumn{2}{|c|}{40~472} &\multicolumn{2}{|c|}{5~000}&\multicolumn{2}{|c|}{5~000} \\
      \hline
      \hline
      & \multicolumn{1}{|c}{Words} & \multicolumn{1}{c|}{Labels} &  \multicolumn{1}{|c}{Words} & \multicolumn{1}{c|}{Labels} &
      \multicolumn{1}{|c}{Words} & \multicolumn{1}{c|}{Labels} \\
      \hline
      \# tokens          & 719~530 & -- & 76~704 & -- & 92~004 & -- \\
      dictionary         &  77~220 &     681 &    15~852 &    501 & 20~149 &     537 \\
      OOV\%   & --     & --     &  30,90  & 0,01  &  37,18  &  0,015  \\
      \hline
    \end{tabular}
    \caption{Statistics of the German corpus TIGER}
  \label{tab:TIGERStats}
  \end{minipage}
\end{table}

We note that the Out-Of-Vocabulary (OOV) rate in the MEDIA corpus is quite small. Indeed there are only two OOV words in both Dev and Test data, moreover they are function words. There is basically no unknown word in this task.
In contrast the OOV rate in both the WSJ and the TIGER corpora is quite high: roughly one word out of five in the first and one out of three in the second.

\subsection{Settings}
\label{subsec:settings}

We used the corpus MEDIA for fine-tuning the hyper-parameters of our models in \cite{DBLP-journals/corr/abs-1904-04733}. MEDIA is the smallest corpus among the three used, and allows thus a faster optimization.
The settings used in this work are the same as those used in \cite{dinarelli_hal-01553830,DBLP-journals/corr/abs-1904-04733}, where the same data were used for evaluation (except for the TIGER corpus).
The settings for the WSJ corpus are the same, except for the word embeddings (\num{300} dimensions) and the learning rate (\num{2.5e-4}).
We use the same settings for WSJ and TIGER. Since  the latter corpus is the largest, we ran few experiments to choose the correct hidden layer size (\num{300}).

Like we already explained in \cite{DBLP-journals/corr/abs-1904-04733}, at first our experiments were not fitting in the memory of our GPUs. In order to solve this problem we found two solutions.
In the first we organize the training data as a single stream of items. This is then split into chunks of fixed size, shifting each chunk by one token with respect to the previous (thus consecutive chunks overlap).
The second solution is more intuitive and consists in clustering sentences of the same length. This way, very long (rare) sentences are assigned to very small clusters, thus fitting in memory, and small and average clusters form bigger clusters, which still fit in memory.
In our experiments we found out that the first solution works far better on the MEDIA corpus.
On the WSJ and TIGER corpora the two solutions are equivalent, we thus prefer the second for these two tasks, which is more intuitive and is expected to work better on bigger corpora.

Our neural network system is coded with pytorch\footnote{https://pytorch.org/docs/stable/index.html}, experiments are run under Linux Debian distribution with GeForce GTX 1080 GPUs.

\subsection{Results}

\begin{table}[t]
\centering
    \begin{tabular}{|l|rcc|}
        \hline
        \textbf{Model}  & \textbf{Accuracy} & \textbf{F1} & \textbf{CER} \\ \hline
        \hline
        \multicolumn{4}{|c|}{\textbf{MEDIA DEV}}\\ \hline
        \hline
        Seq2Biseq \cite{DBLP-journals/corr/abs-1904-04733}				& 89.97	& 86.57	& 10.42 \\
        Seq2Biseq$_{\text{2-opt}}$ \cite{DBLP-journals/corr/abs-1904-04733}	& \textbf{90.22}	& 86.88	& 9.97 \\
        Seq2Biseq-Transformer$_{\text{2-opt}}$     		& 90.08	& 86.93	& 9.87 \\
        Seq2Biseq-Transformer+FEAT$_{\text{2-opt}}$     	& 90.14	& \textbf{87.05}	& \textbf{9.54} \\
        \hline
        \multicolumn{4}{|c|}{\textbf{MEDIA TEST}}\\ \hline
        \hline
        BiGRU+CRF \cite{dinarelli_hal-01553830}				& -- 	& 86.69		& 10.13	\\
        LD-RNN$_{\mathrm{deep}}$ \cite{dinarelli_hal-01553830}	& -- 	& 87.36		& 9.8		\\\hline
        Seq2Biseq \cite{DBLP-journals/corr/abs-1904-04733}				& 89.57	& 87.50	& 10.26	\\
        Seq2Biseq$_{\text{2-opt}}$ \cite{DBLP-journals/corr/abs-1904-04733}	& 89.79	& 87.69	& 9.93 \\
        Seq2Biseq-Transformer$_{\text{2-opt}}$     		& 89.98	& 87.81	& 9.67 \\
        Seq2Biseq-Transformer+FEAT$_{\text{2-opt}}$     	& \textbf{90.12}	& \textbf{87.94}	& \textbf{9.48} \\
        \hline
    \end{tabular}
    \caption{Results of different variants of our model for the semantic tagging task of the corpus MEDIA, compared to the state-of-the-art}
    \label{tab:gru-ldrnn-clen15-vs-SOTA-media}
\end{table}

Results for the MEDIA task are means over \num{10} experiments, the parameters are initialized at random\footnote{Using default initialization of the \emph{Pytorch} library} for each experiment.
On this task we evaluate our models in terms of accuracy and, since the goal is to re-construct semantic units spanning over multiple words, also with F1 measure and \emph{Concept Error Rate} (CER). CER is computed by aligning the model prediction with the gold annotation with the Levenshtein distance algorithm, then the sum of insertions, deletions and substitutions is divided by the length of the gold annotation.
On the other two tasks, since there is no label splitting over multiple words, we evaluate the models only in terms of accuracy.

Since the model presented in this paper is an evolution of the \emph{Seq2Biseq} model proposed in \cite{DBLP-journals/corr/abs-1904-04733}, the model proposed here is named \emph{Seq2Biseq-Transformer}, as it adds characteristics of the \emph{Transformer} on top of the \emph{Seq2Biseq} model.

Results on Dev and Test data of the MEDIA task are shown in table~\ref{tab:gru-ldrnn-clen15-vs-SOTA-media}.
For comparison we show also results presented in our previous works \cite{dinarelli_hal-01553830,DBLP-journals/corr/abs-1904-04733}, which are state-of-the-art on this task.

As described in \cite{DBLP-journals/corr/abs-1904-04733}, the Seq2Biseq model gives better results when trained with \num{2} optimizers (see \cite{DBLP-journals/corr/abs-1904-04733} for details), one for each decoder. The Seq2Biseq-Transformer model is thus always trained with \num{2} optimizers.

As we can see in table~\ref{tab:gru-ldrnn-clen15-vs-SOTA-media}, results obtained on MEDIA with the Seq2Biseq-Transformer model are slightly worse than Seq2Biseq's results in terms of accuracy, and they are slightly better in terms of F1 and CER.\footnote{We did not perform significance tests}
Adding word classes available for the MEDIA task (cf. the begin of section~\ref{sec:eval}, data description) as input to the model, results improve with all evaluation metrics and constitute the new state-of-the-art on this task with manual transcription.\footnote{SLU is performed on automatic transcription, in this work however we evaluate our models against sequence labelling problems}.

\begin{table}[t]
\centering
    \begin{tabular}{|l|c|c|}

        \hline
        \textbf{Model}  & \multicolumn{2}{|c|}{\textbf{Accuracy}} \\ \hline
        \hline
        & WSJ DEV & WSJ TEST \\ \hline
        \hline
        LD-RNN$_{\mathrm{deep}}$ \cite{dinarelli_hal-01553830}		& 96.90	& 96.91 \\
        LSTM+CRF \cite{Ma-Hovy-ACL-2016}						& -- 		& 97.13 \\
        Seq2Biseq \cite{DBLP-journals/corr/abs-1904-04733}			& 97.13	& 97.20 \\
        Seq2Biseq$_{2-opt}$ \cite{DBLP-journals/corr/abs-1904-04733}	& \textbf{97.33}	& 97.35 \\
        Seq2Biseq-Transformer$_{2-opt}$						& \textbf{97.33}	& \textbf{97.36} \\
        \hline
        \multicolumn{3}{|c|}{Models using pre-trained embeddings\footnote{In \cite{DBLP-journals/corr/abs-1902-09113} it is actually not clear if the Transformer and Star-Transformer models have been trained with or without pre-trained embeddings. Authors report ``\emph{Advanced Techniques} except widely-used pre-trained
embeddings'' in the caption of table~5. However results are compared with literature models using pre-trained embeddings, and this is not mentioned in the text.}} \\
        \hline
        Transformer+Char \cite{DBLP-journals/corr/abs-1902-09113}		& -- 		& 97.14 \\
        LSTM+CRF + Glove \cite{Ma-Hovy-ACL-2016}				& 97.46 	& 97.55 \\
        LSTM+LD-RNN + Glove \cite{2018-ijcai-lstm-ldrnn}			& -- 		& 97.59 \\
        Star-Transformer+Char \cite{DBLP-journals/corr/abs-1902-09113}	& -- 	& \textbf{97.64} \\ \hline

    \end{tabular}
\caption{Results of different variants of our model for the POS tagging task on the WSJ corpus, compared to the state-of-the-art}
\label{tab:comp-WSJ}
\end{table}

Results of POS tagging on the WSJ corpus are presented in table~\ref{tab:comp-WSJ}.
As we can see in this case, adding functionalities of the \emph{Transformer} to the Seq2Biseq model give exactly the same result as the Seq2Biseq model on the Dev data, and slightly better on the Test data (on a single experiment).

Keeping both these results and those on the MEDIA task into account, we could conclude that the Seq2Biseq-Transformer model does not really improve our previous model. However we would like to point out that the main goal of this work was to integrate characteristics of the \emph{Transformer} model, and to come up with a generic architecture, which can be applied not only to traditional sequence labelling tasks, but also to tasks such like machine translation and syntactic parsing. Indeed the Seq2Biseq-Transformer architecture has been modified, with respect to our previous model, in order to perform all these tasks. We consider that obtaining similar results (which are still state-of-the-art) with a more versatile model is still a positive result.
Additionally, observing the training procedure of the Seq2Biseq-Transformer model, we found out that while final results are similar, the loss values on both training and Dev data are better than those obtained with the Seq2Biseq model.
This could actually mean the model is over-fitting, and we speculate that instead of using the same settings as in \cite{DBLP-journals/corr/abs-1904-04733}, since the Seq2Biseq-Transformer is more complex (at least in terms of parameters, and see below), a fine-tuning of parameters could lead to better performances.

In table~\ref{tab:comp-WSJ} we can see that, whether we use one or two decoders for training, our models improve the \emph{LSTM+CRF} model \cite{Ma-Hovy-ACL-2016} without using the embeddings pre-trained with \emph{GloVe} \cite{pennington2014glove}.
For a matter of comparison, we show also the best results on this task, obtained with pre-trained embeddings. Though our results are lower than state-of-the-art in terms of accuracy, they are not too far, proving that our hybrid neural architecture is quite effective.
In particular, we compare our results with those reported in \cite{DBLP-journals/corr/abs-1902-09113} obtained with the \emph{Transformer} and \emph{Star-Transformer} models.\footnote{In \cite{DBLP-journals/corr/abs-1902-09113} it is actually not clear if the Transformer and Star-Transformer models have been trained with or without pre-trained embeddings. Authors report ``\emph{Advanced Techniques} except widely-used pre-trained
embeddings'' in the caption of table~5. However results are compared with literature models using pre-trained embeddings, and this is not mentioned in the text.}
Whether the \emph{Transformer} model has been trained with or without pre-trained embeddings, our models outperform it by a significant margin. As we mentioned in section~\ref{subsec:bestof3worlds}, we have experienced a drop of 3 points in the F1 measure using a \emph{Transformer} for mention detection in preliminary experiments (not reported in this work). We hypothesize that the one-to-one correspondence between input and output units in sequence labelling tasks is a very useful information, and the lack of performance of the \emph{Transformer} on sequence labelling tasks can be explained by the fact that this model doesn't use it.

\begin{table}[t]
    \centering
    \begin{tabular}{|l|c|c|}

        \hline
        \textbf{Model}  & \multicolumn{2}{|c|}{\textbf{Accuracy}} \\ \hline
        \hline
        & TIGER DEV & TIGER TEST \\ \hline
        \hline
        Seq2Biseq$_{2-opt}$						& \textbf{93.90} (98.30)	& \textbf{91.86} (97.74) \\
        \hline
        \hline
        VO-CRF \cite{Lavergne17learning}	& -- 		& 88.78 \\ \hline
    \end{tabular}
    \caption{Results of different variants of our neural architecture for the morpho-syntactic tagging task on the TIGER corpus, compared with state-of-the-art}
\label{tab:comp-TIGER}
\end{table}

In table~\ref{tab:comp-TIGER} we show results obtained on the morpho-syntactic tagging task on the German corpus TIGER.
To the best of our knowledge, the best results in the literature for this task are reported in \citet{Lavergne17learning}, obtained using a \emph{Variable Order} CRF (VO-CRF).
We can see thus that our model improves the state-of-the-art on this task. It is important to note however that the TIGER corpus has been less used for the evaluation of neural models for sequence labelling. Results reported in \citet{Lavergne17learning} are indeed obtained with a discrete-representation based, though quite sophisticated, CRF model.
In table~\ref{tab:comp-TIGER} we report also, between parenthesis, POS tagging results. These are obtained from morpho-syntactic tagging results extracting the POS tag from the complete label. That is no model training has been performed for the POS tagging task alone.

\section{Conclusions}\label{sec:conclusions}

In this paper we proposed a neural model for sequence labelling integrating characteristics of the most successful models of the last couple of years: bidirectional RNNs, \emph{Sequence-to-Sequence} model, and the \emph{Transfomer} model.
An evaluation on three tasks of sequence labelling shows that the proposed model is very effective for this kind of tasks.
Indeed it achieves results comparable to the state-of-the-art, but keeping a generic architecture that allows to use the same model on other kind of tasks like machine translation and syntactic parsing. Moreover in some cases our model outperforms state-of-the-art models, proving that our hybrid architecture has its place among current neural architectures.

\section*{Acknowledgments}

This research is part of the program ``Investissements d’Avenir'' funded by the French National Research Agency (ANR) ANR-10-LABX-0083 (Labex EFL).

This work has been partially funded by the French ANR project DEMOCRAT (\emph{``Description et modélisation des chaînes de référence : outils pour l'annotation de corpus et le traitement automatique''}) project ANR-15-CE38-0008.

\renewcommand\refname{References}
\bibliographystyle{taln2019}
\bibliography{talnbiblio}

\end{document}